\documentclass{article}

\usepackage{arxiv}

\usepackage[utf8]{inputenc} 
\usepackage[T1]{fontenc}    
\usepackage{hyperref}       
\usepackage{url}            
\usepackage{booktabs}       
\usepackage{amsfonts}       
\usepackage{nicefrac}       
\usepackage{microtype}      
\usepackage{lipsum}		
\usepackage{graphicx}
\usepackage{amsmath}
\usepackage{doi}
\usepackage{comment}
\usepackage{caption}
\usepackage{cite}
\usepackage{subcaption}
\usepackage{authblk}
\usepackage[english]{babel}
\usepackage{csquotes}

\usepackage{subcaption}
\usepackage{caption}
\usepackage{xcolor}

\usepackage{pgfplots}
\usepackage{pgfplotstable}
\pgfplotsset{compat=1.17}
\usepackage{filecontents} 
\usepackage{pgfplots}
\usepgfplotslibrary{colorbrewer}
\pgfplotsset{compat = 1.15, cycle list/Set1-8} 
\usetikzlibrary{pgfplots.statistics, pgfplots.colorbrewer} 
\usepackage{pgfplotstable}
\usepackage{filecontents}

\newcommand{\bs}[1]{\boldsymbol{#1}}

\title{A Graph Neural Network approach to zero-shot Digital Twins}

\renewcommand{\shorttitle}{Graph neural networks informed locally by thermodynamics}

\hypersetup{
pdftitle={A Graph Neural Network approach to zero-shot Digital Twins},
pdfsubject={q-bio.NC, q-bio.QM},
pdfauthor={A. Tierz},
pdfkeywords={First keyword, Second keyword, More},
}
\author[1]{Alicia Tierz}
\author[1]{Icíar Alfaro}
\author[1]{David González}
\author[1]{Elías Cueto}
\affil[1]{{\small Keysight-UZ Chair of the Spanish National Strategy on  Artificial Intelligence. \protect\\ Aragon Institute of Engineering Research (I3A). Universidad de Zaragoza. Zaragoza, Spain.}}

\begin{document}
\maketitle

\begin{abstract}
Traditional Predictive Digital Twins often remain geometrically rigid, requiring extensive retraining or fine-tuning whenever the underlying physical domain or boundary conditions change. To overcome this limitation, we present a novel framework for \textit{Zero-Shot Digital Twins} that seamlessly couples real-time visual perception with a geometry-agnostic, physics-informed reasoning engine. At the core of our architecture is the Thermodynamics-Informed Graph Neural Network architecture, a Geometric Deep Learning solver grounded in a metriplectic thermodynamic formalism that enforces energy conservation and non-negative entropy production locally through graph message passing. The framework integrates an auxiliary Graph Neural Network to infer unobservable fields (such as stress tensors or velocity and energy distributions) directly from sparse initial visual boundaries, mitigating numerical start-up transients. To bridge the sim-to-real gap, we implement a continuous closed-loop data assimilation mechanism; the pipeline tracks macroscopic deformations and free-surface fluid boundaries in real-time using deep segmentation networks combined with sparse optical flow, dynamically correcting the autoregressive simulation rollout and eliminating numerical drift. To test the validity of our approach, we demonstrate the extreme generalization capabilities of our approach across two disparate physical regimes: the large deformations of a viscoelastic beam and the non-linear sloshing of a viscous fluid. In both scenarios, the unified framework instantiates physically accurate simulations on novel, unseen geometries without case-specific retraining, operating well within real-time latency budgets (approximately $25$~ms per frame) and enabling the direct projection of latent mechanical variables via Augmented Reality.
\end{abstract}

\section{Introduction}

Computational simulation has served as a cornerstone across scientific disciplines for decades, facilitating the prediction of physical phenomena and enabling the refinement of engineering designs before costly physical prototypes are pursued. Traditionally, these simulations relied heavily on mechanistic governing equations expressed through partial differential equations (PDEs) to model complex systems in fields like fluid dynamics or structural mechanics. However, the onset of the information era, often referred to as the fourth paradigm of science \cite{Tolle:2011}, has driven a shift toward data-driven approaches. Deep learning, in particular, offers solutions for handling complex non-linear dynamics under strict real-time constraints, a necessity for modern applications such as predictive Digital Twins \cite{tao2019digital}.

While data-driven models, or learned simulators, offer the benefit of rapid inference compared to traditional methods like the Finite Element Method (FEM) or Computational Fluid Dynamics (CFD), they frequently present significant challenges. Deep learning algorithms are computationally demanding during training, data-hungry, and critically struggle with scalability and generalization. Purely data-based ``black-box'' networks lack explicit physical interpretability and robustness, often resulting in unreliable predictions when faced with out-of-distribution (OOD) conditions. To ensure trust in predictions, embedding physical principles into machine learning models has become essential.

To address these limitations, recent research in Geometric Deep Learning (GDL) has successfully leveraged inherent problem structures to create robust simulators \cite{bronstein2021geometricdeeplearninggrids, battaglia2018relational}. Notably, the development of Thermodynamics-Informed Neural Networks (TINNs) and their Graph Neural Network counterpart, TIGNNs, \cite{TIERZ2025110108}, has provided a solution that employs the GENERIC formalism \cite{Grmela1997} to enforce energy conservation and entropy production. Unlike earlier architectures that relied on costly global matrix assembly, the Local-TIGNN preserves the efficient node-by-node structure of Graph Neural Networks (GNNs) popularized by state-of-the-art learned physical simulators \cite{sanchez2020learning, pfaff2021learning}, offering a scalable engine capable of adhering to the first and second laws of thermodynamics.

However, possessing a fast and physically consistent simulator is only one component of a truly autonomous system. Current Digital Twin implementations often remain rigid, requiring extensive retraining or fine-tuning whenever the underlying geometry or physical context changes \cite{jinzhi2022exploring}. This limitation hinders the deployment of Digital Twins in dynamic, unstructured environments where the system must adapt to novel objects instantaneously.

In this paper, we leverage the robust foundation of the Local-TIGNN to present a novel framework for predictive perception through the development of \textbf{Zero-Shot Digital Twins} of previously unseen objects in the scene. By ``predictive perception'' we mean a type of perception that is not only quantitative---in contrast to human perception---but also enables predictions to be made about quantities that govern the physics of the scene, such as stresses or energy, which remain hidden from the human eye.

We propose a comprehensive architecture that transcends traditional offline simulation by integrating the Local-TIGNN engine with a real-time visual perception pipeline. This cognitive framework dynamically reconstructs the geometry of objects never encountered during training, such as fluid containers or highly deformable solids, and immediately instantiates a physics-informed simulation without the need for model retraining.

Crucially, our proposed framework incorporates a continuous closed-loop feedback mechanism that utilizes real-time visual tracking to correct the predicted physical state online. By dynamically nudging the autoregressive rollout, this assimilation step compensates for accumulated time integration errors and ensures strict alignment with the observed physical reality. Finally, the system seamlessly projects computationally derived, unobservable state variables (such as internal energy or stress) directly onto the physical scene via Augmented Reality (AR). This pipeline effectively transforms the Digital Twin from a rigid, offline predictive tool into a zero-shot augmented intelligence system, capable of reasoning about and visualizing complex physical dynamics on the fly.

The remainder of this paper is structured as follows. Section II reviews related work in physics-informed modeling, geometric deep learning, and cognitive digital twins. Section III details the methodology, explaining the architecture of the Local-TIGNN and the mechanism for the real-time perception and prediction loop. Section IV discusses the demonstrated capacity for generalization and AR visualization. Finally, Section V presents numerical experiments and validation, followed by Section VI, which offers conclusions and directions for future work.

\section{Related Work}

\subsection{Physics-Informed Deep Learning}

The simulation of complex physical phenomena has traditionally relied on rigorous numerical methods derived from conservation laws, expressed through Partial Differential Equations (PDEs). While accurate, these solvers are often computationally expensive for real-time applications. The advent of data-driven modeling and simulation sought to overcome these costs; however, early deep learning architectures typically lacked robustness and physical interpretability, struggling significantly with generalization when exposed to out-of-distribution (OOD) regimes \cite{willard2022}.

To address these shortcomings, the paradigm of Physics-Informed Deep Learning (PIDL) emerged, aiming to embed scientific priors directly into learning algorithms \cite{karniadakis2021physics}. Physics-Informed Neural Networks (PINNs), incorporate governing PDEs as soft constraints within the loss function \cite{RAISSI2019}. Nevertheless, PINNs generally require explicit knowledge of the governing equations and continuous retraining for new boundary conditions, limiting their flexibility in dynamic environments where equations may be unknown or parameters vary rapidly.

Initial efforts to enforce compliance to physical laws in learned models focused on conservative systems. Architectures based on Lagrangian and Hamiltonian formalisms were developed to guarantee symplectic structure and energy conservation \cite{greydanus2019hamiltonian, cranmer2020discovering}. However, real-world engineering systems, such as fluid dynamics and solid mechanics, are inherently dissipative and irreversible, dynamics that purely conservative frameworks cannot capture. Modeling such systems requires strict adherence to both the first and second laws of thermodynamics.

To integrate these principles, recent methodologies have adopted metriplectic formalisms \cite{MORRISON1986}, specifically the General Equation for Non-Equilibrium Reversible-Irreversible Coupling (GENERIC) \cite{Grmela1997}. Models leveraging this framework, often termed Structure-Preserving Neural Networks (SPNNs) or Thermodynamics-Informed Neural Networks (TINNs), guarantee energy conservation and non-negative entropy production by construction \cite{HERNANDEZ2021109950, hernandez2022thermodynamics}. While early SPNNs proved effective for low-dimensional systems, scaling them to large, unstructured meshes revealed a critical bottleneck: the reliance on global Poisson ($\bs L$) and dissipation ($\bs M$) matrices. This global dependency breaks the local message-passing structure of Graph Neural Networks (GNNs), leading to quadratic memory complexity and impeding scalability \cite{Geiger_2020}.

Addressing this scalability challenge, recent work has bridged geometric deep learning with localized thermodynamic constraints. This effort culminated in the development of the Local-TIGNN (Thermodynamics-Informed Graph Neural Network) architecture \cite{hernandez2023port, TIERZ2025110108}. By adopting a port-metriplectic perspective, this framework treats each node in the graph as an open thermodynamic system that exchanges energy and entropy fluxes through ``ports'' with its neighbors. This formulation avoids the assembly of global matrices, recovering the linear computational efficiency of GNNs while strictly enforcing thermodynamic laws. This local, scalable engine serves as the foundational backbone for the Zero-Shot Cognitive Digital Twin strategy presented in this work.

\subsection{Graph Neural Networks and Geometric Biases}

Deep learning on Euclidean domains (e.g., images) faces inherent limitations when applied to the unstructured, irregular discretizations typical of physical simulations. To overcome these challenges, Geometric Deep Learning (GDL) generalizes neural networks to non-Euclidean domains by leveraging symmetries and geometric priors \cite{bronstein2021geometricdeeplearninggrids}. Within this framework, Graph Neural Networks (GNNs) \cite{Scarselli_GNN} have emerged as the standard for modeling complex topologies, offering a flexible alternative to grid-based convolutional approaches.

In the context of physical simulation, GNNs operate on the principle of Message Passing (MPNNs) \cite{pmlr-v70-gilmer17a}. By iteratively propagating latent information across local neighborhoods, these architectures learn to approximate complex differential operators directly on the mesh. Prominent models like MeshGraphNets have successfully employed this Encoder-Processor-Decoder scheme to simulate fluid dynamics and structural mechanics with high fidelity \cite{pfaff2021learning, tierz2024feasibilityfoundationalmodelssimulation}. 

However, standard GNN simulators often operate as purely data-driven black boxes. While efficient, they lack explicit physical interpretability and struggle to strictly adhere to conservation laws, especially in low-data regimes. To bridge this gap, the geometric flexibility of GNNs must be coupled with the thermodynamic rigor discussed in the previous section. By adopting the Local-TIGNN architecture \cite{TIERZ2025110108}, we effectively embed the GENERIC formalism into the message-passing updates. This integration transforms the GNN from a statistical approximator into a robust, structure-preserving solver, providing the necessary foundation to build a Zero-Shot Digital Twin capable of operating on unseen geometries.

\subsection{Digital Twins, Perception, and Augmented Reality}

The demand for predictive reliability in modern engineering has spurred the evolution of Digital Twins (DTs), virtual replicas designed to continuously evolve in synchronization with their physical counterparts \cite{Grieves2017, Kapteyn2020}. While traditional DTs often rely on pre-defined models, the field is shifting toward Cognitive Digital Twins, CDTs, \cite{moya2022digital, chinesta2020virtual}. Unlike standard predictive tools, CDTs act as augmented intelligence systems capable of interpreting the current dynamic state of a system to reason about its future evolution, often requiring the integration of semantic knowledge and advanced reasoning capabilities \cite{d2022cognitive}.

A critical prerequisite for robust CDTs is physical scene understanding, or the ability to perceive and reconstruct the environment in real-time. Traditional data-driven simulators struggle with generalization when faced with geometries outside their training set. To overcome this, recent frameworks integrate computer vision pipelines directly with physics engines. By dynamically generating mesh representations from visual sensors (e.g., RGB-D cameras), these systems can instantiate simulations for objects never encountered during training, enabling a ``Zero-Shot'' deployment capability \cite{Xian2017, liu2023digital}. Furthermore, because real-world sensor data is often sparse or noisy, frequently capturing only boundary deformations or the free surface of a fluid, purely open-loop predictions quickly degrade. To address this, the paradigm of Hybrid Twins \cite{chinesta2018} integrates continuous closed-loop feedback mechanisms. Historically, these systems have relied on established data assimilation techniques \cite{cheng2023machine} to correct the predicted physical state using real-time observations, thereby preventing numerical drift and ensuring alignment with reality \cite{Moya2020}. However, existing data assimilation strategies are largely constrained to predefined, geometrically static physical models. Applying continuous, vision-based correction mechanisms to geometry-agnostic learned simulators to enable true zero-shot deployment on unobserved geometries without retraining remains a significant open challenge that our proposed framework directly addresses.

The ultimate utility of a Cognitive DT lies in its interface with the human operator. To this end, Augmented Reality (AR) has emerged as a fundamental technology for Intelligence Augmentation (IA). AR transcends passive visualization by seamlessly blending virtual information with the physical environment \cite{milgram1994taxonomy}. By projecting computationally derived variables, such as internal energy, velocity fields, or stress tensors, 
directly onto the observed physical object, AR renders invisible physical quantities perceptible to the naked eye. This capability transforms the Digital Twin from a background computational engine into an interactive decision-support tool, enhancing human understanding in complex industrial and engineering scenarios \cite{fraga2018review}.

\section{Methodology: Zero-Shot Cognitive Framework}

This section details the architectural integration of the thermodynamic inference engine with the real-time perception pipeline. The proposed framework operates as a closed-loop system comprising three modular stages: (1) the \textit{Local-TIGNN} solver, which serves as the generalized physics engine; (2) the \textit{Visual Perception Module}, which reconstructs novel geometries from camera inputs; and (3) the \textit{Data Assimilation Loop}, which synchronizes the simulation with reality to correct numerical drift.

\subsection{The Local-TIGNN Engine: A Geometry-agnostic Solver}

The predictive core of our framework is the Local-TIGNN \cite{TIERZ2025110108}, a geometric deep learning architecture designed to function as a generalized physics engine. To ensure that the predicted dynamics are physically robust and trustworthy, even when applied to unseen geometries, the solver is grounded in the General Equation for Non-Equilibrium Reversible-Irreversible Coupling (GENERIC) formalism \cite{Grmela1997}.

\subsubsection*{Thermodynamic Foundation (Global GENERIC)}
The GENERIC formalism describes the time evolution of an isolated system's state variables $\bs z$ by coupling reversible and irreversible dynamics. The general evolution equation is given by:
\begin{equation}\label{GENERIC2}
    \dot{ \bs z} = \bs L( \bs z) \frac{\partial E}{\partial \bs z} + \bs M( \bs z)\frac{\partial S}{\partial \bs z},
\end{equation}
where $E(\bs z)$ and $S(\bs z)$ represent the total energy and entropy of the system, respectively. The geometric structure of the dynamics is defined by the Poisson matrix $\bs L(\bs z)$, which must be skew-symmetric and governs the reversible portion, and the friction matrix $\bs M(\bs z)$, which must be symmetric and positive semi-definite to ensure non-negative entropy production and governs dissipation.

To ensure adherence to the first and second laws of thermodynamics, the formulation imposes the so-called degeneracy conditions:
\begin{equation}\label{deg1}
    \bs L(\bs z)\frac{\partial S}{\partial \bs z} = \bs 0,  \quad
    \bs M(\bs z)\frac{\partial E}{\partial \bs z} = \bs 0.
\end{equation}
The first equation ensures that reversible dynamics do not produce entropy, while the second equation  guarantees that dissipative processes conserve the total energy of the system.

\subsubsection*{Nodal Port-Metriplectic Formulation}
While Equation (\ref{GENERIC2}) is rigorous, constructing global $\bs L$ and $\bs M$ matrices is computationally prohibitive for large systems and incompatible with the (mostly) local nature of Graph Neural Networks. To address this, we employ a Nodal Port-Metriplectic formulation. 

The system is discretized as a graph $\mathcal{G}=(\mathcal{V}, \mathcal{E})$, where the global dynamics are decomposed into local nodal contributions. Each node $i$ acts as an open thermodynamic subsystem that interacts with its neighbors $\mathcal{N}(i)$ through ports (the edges of the graph). The state evolution of a single node $i$ is governed by:
\begin{equation}\label{generic2}
\dot{\bs z}_i  = \bs L_i (\bs z_i)  \frac{\partial e_i}{\partial \bs z_i} + \bs M_i(\bs z_i)\frac{\partial s_i}{\partial \bs z_i}  - \sum_{j \in \mathcal{N}(i)} \left[ \bs L_{ij} (\bs z_j)\frac{\partial e_j}{\partial \bs z_j} + \bs M_{ij}(\bs z_j) \frac{\partial s_{j}}{\partial \bs z_j} \right],
\end{equation}

where the terms involving index $i$ define the bulk dynamics of the open thermodynamic subsystem (node i). Here, the matrices $\bs L_i$ and $\bs M_i$ represent the self-dynamics of the node, while $\bs L_{ij}$ and $\bs M_{ij}$ represent the flux interactions exchanged with neighbour $j$.

The degeneracy conditions remain the same but are applied at the particle level. 
\begin{equation}\label{deg2}
\bs L_{\text{bulk}}(\bs z)\frac{\partial S_{\text{bulk}}}{\partial \bs z} = \bs 0,  \quad
\bs M_{\text{bulk}}(\bs z)\frac{\partial E_{\text{bulk}}}{\partial \bs z} = \bs 0.
\end{equation}

These constraints ensure the bulk dynamics conserve the local energy $(e_i)$ and that the energy potential does not affect entropy production.

\subsubsection*{Implementation via Graph Neural Networks}
This nodal formulation aligns perfectly with the Message Passing mechanism of GNNs. The summation term in Eq. (\ref{generic2}) corresponds to the aggregation step in a GNN, where the network learns to approximate the interaction matrices ($\bs L_{ij}, \bs M_{ij}$) based on local edge features. By enforcing the skew-symmetry of $\bs L$ and the positive semi-definiteness of $\bs M$ locally, the Local-TIGNN guarantees thermodynamic consistency by construction. Furthermore, by incorporating the degeneracy conditions (\ref{deg2}) as a soft constraint in the loss function, the model is strictly compelled to satisfy the existence of stable equilibrium states. Crucially, because the physics are learned as local interaction rules rather than global mappings, the trained model is geometry-agnostic and can be deployed zero-shot on novel meshes generated by the perception system.

\subsection{Real-Time Perception and Graph Generation}

The cognitive ability of the Digital Twin relies on a computer vision pipeline capable of interpreting the physical scene in real-time. The system receives a raw video stream from a standard RGB camera. To handle different material behaviors, we implement distinct reconstruction strategies for solid and fluid objects.

\subsubsection{Graph Construction for Solids}

For deformable solids, defining a consistent Lagrangian description of kinematics is essential for tracking deformation stresses. However, homogeneous surfaces often lack distinct visual features required for robust motion tracking. To overcome this, we apply a high-contrast fiducial grid pattern to the object's surface. This pattern acts as a physical texture map, facilitating the extraction of a structured graph topology. This is, however, by no means mandatory. Previous works by the authors did not employ fiducial markers, see, for instance, \cite{badias2021morph}. For solids, the connectivity matrix of the graph is fixed throughout the computation.

The reconstruction process operates in two stages: \textit{Initialization} and \textit{Temporal Tracking}.

\paragraph*{\textbf{Topology Initialization (Keyframe $t_0$)}}
In the initial frame, the system employs a semantic segmentation network, more specifically, the U-net architecture is utilized to identify the grid structure \cite{ronneberger2015u}. The network processes the RGB input and outputs a binary mask separating the grid lines from the background.

From this segmentation mask, a skeletonization procedure is performed to extract the centerlines of the grid. To formalize the graph structure, the nodes $\mathcal{V}$ are identified by detecting the intersection points within the skeletonized mask. This is achieved through the Shi-Tomasi corner detection algorithm \cite{shi1994good}, which identifies points of high intensity variance in multiple directions—corresponding to the grid crossings. Specifically, the method calculates the minimum eigenvalue of the spatial gradient matrix at each pixel, selecting coordinates where this value exceeds a predefined threshold and satisfies a minimum distance criterion between neighboring detections. This ensures a robust, one-to-one mapping with the physical grid intersections. Finally, the edges $\mathcal{E}$ are established based on the skeletal connectivity between these nodes, defining the reference topology $\mathcal{G}_0$.

\paragraph*{\textbf{Real-Time Temporal Tracking ($t > 0$)}}
Running the semantic segmentation and skeletonization pipeline for every frame is computationally expensive and can introduce temporal jitter due to segmentation noise. To ensure real-time performance and temporal coherence, we transition to a tracking-based approach after initialization.

We utilize the Lucas-Kanade method with pyramidal implementation \cite{lucas1981iterative} (Sparse Optical Flow) to track the pixel coordinates $(u, v)$ of the identified nodes across subsequent frames. This algorithm assumes brightness constancy and spatial coherence to estimate the motion vector of each node with sub-pixel accuracy.

\paragraph*{\textbf{3D World Projection}}
Finally, the tracked 2D pixel coordinates $(u, v)$ are mapped into the 3D world space $(X, Y, Z)$. Since a single RGB camera is used, the depth information $Z$ is not directly measured but is instead assumed to be a known constant $d$, corresponding to the physical distance between the camera plane and the structural arrangement. Using the camera's intrinsic matrix $\bs{K}$, the 3D coordinates are reconstructed as follows:
\begin{equation}
    \begin{bmatrix} X \\ Y \\ Z \end{bmatrix} = d \cdot \bs{K}^{-1} \begin{bmatrix} u \\ v \\ 1 \end{bmatrix}.
    \label{eq:monocular_projection}
\end{equation}
By leveraging this geometric constraint, we obtain a dynamic 3D graph $\mathcal{G}_t$ that captures the instantaneous deformation of the beam. This representation is subsequently fed into the Local-TIGNN engine, providing the necessary boundary conditions or serving as input for data assimilation.

\subsubsection{Graph Construction for Fluids}

Unlike deformable solids, we employ for (free-surface) fluids an updated Lagrangian kinematic description where the graph topology is not fixed but evolves dynamically due to particle motion. Consequently, the reconstruction pipeline cannot rely on a persistent grid. Instead, we generate a point cloud representation that adapts to the fluid's deformation. The process follows a parallel structure to the solid case, comprised of: \textit{Initialization}, \textit{Surface Tracking}, and \textit{3D Projection}.

\paragraph*{\textbf{Geometry Initialization and State Inference (Keyframe $t_0$)}}
Similar to the solid case, the initialization is driven by semantic perception via a YOLO-seg architecture. The network processes the RGB input to segment two critical classes: the container and the fluid. The system monitors the container's velocity to detect the precise instance when external excitation ceases, establishing this timestamp as $t_0$.

Extracting a volumetric 3D particle mesh from a single monocular viewpoint under highly dynamic, asymmetric sloshing conditions represents a significant geometric challenge. To overcome this without assuming strict rotational symmetry at the free surface, a robust three-step reconstruction procedure is implemented. First, the 2D liquid free surface is extracted from the boundary of the segmentation mask, and the absolute maximum fluid height is identified. Second, a 3D bounding volume representing the interior geometry of the container up to this maximum height is instantiated and populated via Poisson disk sampling \cite{bridson2007fast}, ensuring a uniform internal node density. Third, a vertical binning and trimming operation is executed: the 3D point cloud is discretized into a localized grid of vertical columns. For each column, the true fluid elevation is retrieved by projecting the corresponding region of the 2D semantic mask into the 3D space, and any generated particles lying above this localized threshold are clipped. Finally, the graph connectivity $\mathcal{E}$ is established dynamically via a distance search with a cutoff distance $r_c$, defining the interaction ports required by the metriplectic formalism.

\paragraph*{\textbf{Real-Time Surface Tracking ($t > 0$)}}
While the internal particle motion is predicted by the Local-TIGNN engine, the vision pipeline provides real-time boundary observations to constrain the simulation. Given the chaotic nature of fluid flows, tracking individual Lagrangian particles is visually infeasible. Instead, we implement a contour-tracking algorithm to extract the free-surface profile $h_{\text{target}}(\bs{x})$ at each time step.

To assimilate this visual data without disrupting internal dynamics, a column-wise vertical rescaling strategy is formulated. For every spatial column $\bs x$, a scaling factor $\gamma(\bs x)$ quantifies the discrepancy between the observed boundary and the simulated state:
\begin{equation}
    \gamma(\bs x) = \frac{h_{\text{target}}(\bs x) - h_{\text{bottom}}}{h_{\text{pred}}(\bs x) - h_{\text{bottom}}}.
\end{equation}
Crucially, this factor is applied across the entire vertical coordinate of the fluid column rather than exclusively at the surface:
\begin{equation}
    z_{\text{new}}^{(i)} = h_{\text{bottom}} + \gamma(\bs x_i) \cdot (z^{(i)} - h_{\text{bottom}}),
\end{equation}
where $z^{(i)}$ denotes the vertical position of particle $i$. This operation acts as a continuous volumetric deformation mapping that aligns the simulated fluid volume with the observed surface geometry. By rescaling the entire column, the internal relative distribution of particles is preserved, and local volume consistency is maintained, thereby preventing artificial clustering or void formations typical of naive surface-only nudging. Finally, to ensure numerical stability and prevent overshoot, we impose $z_{\text{final}}^{(i)} = \min(z_{\text{new}}^{(i)}, \, h_{\text{target}}(\bs x_i))$.

\paragraph*{\textbf{3D World Projection}}
Finally, the 2D coordinates of the generated particles and the tracked surface profile are mapped into the 3D world space. Consistent with the solid experiment, we assume the fluid motion is primarily planar and parallel to the image plane at a known depth $d$.

Using the camera's intrinsic matrix $\bs{K}$, the 2D pixel coordinates $(u, v)$ of the fluid particles are reconstructed into physical coordinates $(X, Y, Z)$ using Eq. (\ref{eq:monocular_projection}). This geometric transformation anchors the virtual fluid domain to the physical container, ensuring that the Digital Twin operates in a metric space consistent with the real-world experiment.

\subsection{Inference of Latent Fields and State Initialization}
\label{subsec:inference_hidden_fields}

The Local-TIGNN architecture performs physical reasoning by reconstructing the full thermodynamic state $\bs{z}$ from sparse, observable boundary data $\partial \Omega_{\text{obs}}$. This represents a fundamental inverse problem: while the computer vision system provides real-time boundary geometry, the internal state remains strictly latent. To bridge this gap, we introduce an auxiliary initialization network, $\Psi_{\text{ini}}$, which maps observable nodal coordinates $\bs{q}$ (and, where applicable, auxiliary kinematic inputs) to the unobserved internal state variables. Rather than acting as an unconstrained black-box regressor, the inference performed by $\Psi_{\text{ini}}$ is safeguarded by the mathematical structure of the GENERIC formalism. This ensures that the generated initial fields are thermodynamically consistent by construction, respecting both energy conservation and entropy inequality constraints. Consequently, this architecture effectively resolves the ``cold start'' problem typical of vision-based digital twins \cite{moya2023thermodynamics, jinzhi2022exploring}, bypassing the severe numerical transients or unphysical shocks inherent in assuming null or uncalibrated initial conditions.

This initialization is tailored to the specific physics of each domain:
\begin{itemize}
    \item \textbf{Deformable Solids (Static Initialization):} For deformable structures, assuming a stress-free initial state contradicts the static equilibrium under self-weight. Here, $\Psi_{\text{ini}}$ acts as a learned static solver, mapping deformed nodal positions $\bs{q}$ to the initial stress tensor $\boldsymbol{\sigma}$. By pre-computing this equilibrium field, the engine accurately identifies internal stress hotspots without requiring a multi-step numerical settling process.
    \item \textbf{Fluid Dynamics (Dynamic Initialization):} In the case of free-surface flows within moving containers, the initial state is defined by the fluid's volume geometry and the container's velocity. Given the fluid's mass and the kinematic history of the vessel, the module infers the internal velocity field $\bs{v}(\bs{x}, t)$ and the distribution of internal energy density $\bs{e}(\bs{x}, t)$. This reconstructs the momentum and dissipation patterns necessary to predict the subsequent sloshing dynamics accurately.
\end{itemize}
By decoupling these initialization strategies under a unified thermodynamic framework, $\Psi_{\text{ini}}$ provides a rigorous foundation for real-time inference, allowing the simulation to proceed seamlessly from perceived geometry to latent physical reasoning.

\subsection{Closed-Loop Data Assimilation}

Open-loop simulations inevitably diverge from reality over time due to accumulated numerical integration errors and unmodeled external disturbances. To address this, we implement a \textit{Hybrid Twin} strategy characterized by a continuous feedback loop that integrates the physical priors with real-time visual observations \cite{chinesta2018}.

At each time step $\Delta t$, the system performs a prediction-correction cycle:
\begin{enumerate}
    \item \textbf{Prediction Step (Physics):} The Local-TIGNN predicts the state evolution $\hat{\bs{z}}_{t+1}$ based on the learned thermodynamic operators.
    
    \item \textbf{Observation Step (Perception):} The computer vision pipeline captures the instantaneous configuration of the observable domain boundaries $\bs{q}_{\text{obs}, t+1}$. In the structural case, these correspond to the tracked fiducial nodes, whereas for the fluid, $\bs{q}_{\text{obs}}$ represents the discretized free surface profile extracted from the segmentation mask.
    
    \item \textbf{Correction Step (Assimilation):} The predicted positions are effectively ``nudged'' toward the observed reality via a Newtonian relaxation formalism \cite{hoke1976initialization, asch2016data}. This acts as a boundary constraint—forcing the solid's geometry or the fluid's top-layer particles to match the video feed. This ensures that the simulation remains anchored to the physical world while the neural network infers the \textit{hidden} state variables (e.g., stress tensors for the solid; internal velocity fields and internal energy distributions for the fluid) that the camera cannot directly see.
\end{enumerate}

This synchronization ensures that the Augmented Reality projection, which visualizes these hidden fields, remains spatially coherent with the real object, providing the user with a reliable, physics-informed view of the system's internal behavior.

\begin{figure*}[htbp]
    \centering
    \begin{subfigure}[b]{\textwidth}
        \centering
        \includegraphics[width=\textwidth]{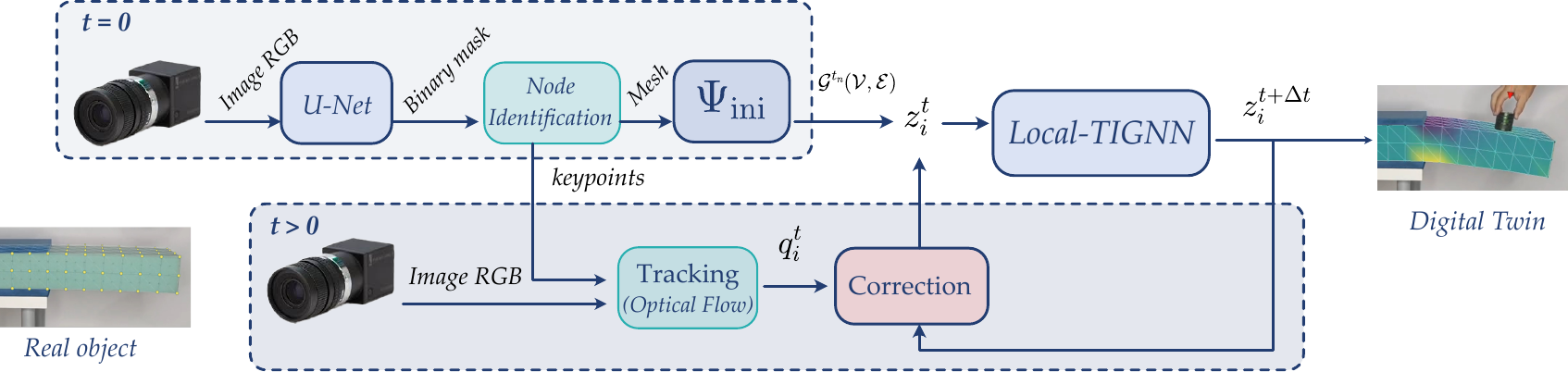}
        \caption{} 
        \label{fig:beam_pipeline}
    \end{subfigure}
    
    \par\bigskip 

    \begin{subfigure}[b]{\textwidth}
        \centering
        \includegraphics[width=\textwidth]{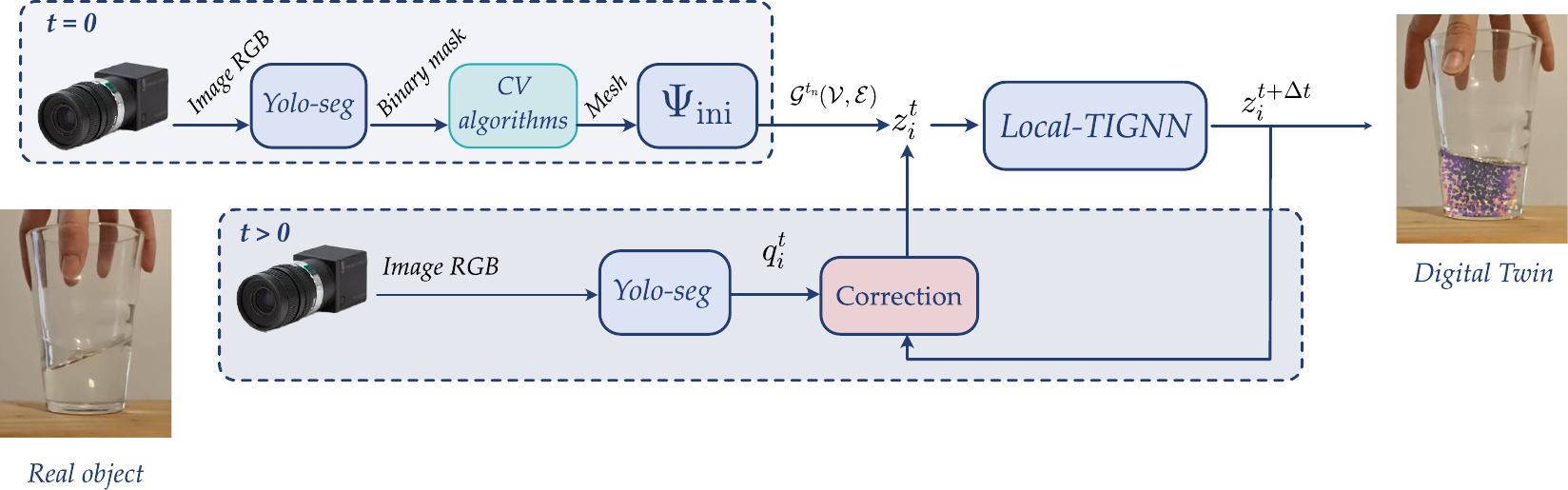}
        \caption{} 
        \label{fig:glass_pipeline}
    \end{subfigure}
    
    \caption {Comparison of Digital Twin Pipelines. 
(a) In the solid mechanics scenario, semantic segmentation (U-Net) and node identification are executed exclusively at $t=0$ to establish the initial mesh. An auxiliary initialization GNN ($\Psi_{\text{ini}}$) is then deployed to infer the initial internal stress state. For subsequent steps ($t>0$), the pipeline capitalizes on Optical Flow to track grid intersections for real-time state correction. 
(b) In the fluid dynamics scenario, the lack of trackable surface features necessitates continuous semantic segmentation (YOLO-seg) across all time steps ($t \ge 0$). At $t=0$, the initialization GNN ($\Psi_{\text{ini}}$) infers the non-zero latent state variables (velocity fields and internal energy distributions) from the initial geometry, establishing a physically consistent state $z_i^t$ for the Local-TIGNN rollout.}
    \label{fig:optimization_strategy}
\end{figure*}

\section{Cognitive Digital Twins and Generalization}
\label{sec:cognitive_dt}

The transition from traditional numerical simulation to a Cognitive Digital Twin (CDT) represents a foundational shift from static replication to dynamic interpretation. This work frames the Digital Twin not only as a high-fidelity mirror of a physical asset, but as an active system endowed with physical scene understanding. Unlike conventional twins that rely on rigid, predefined spatial discretizations, the proposed framework aligns with the paradigm of augmented intelligence \cite{d2022cognitive}. Within this scope, the underlying model serves as a physical reasoning engine capable of interpreting instantaneous dynamic states to infer future evolution, even when subjected to boundary conditions and geometries entirely absent from the training set.

\subsection{Zero-Shot Generalization via Local Learning}
The core methodological contribution of this framework lies in its capacity to achieve zero-shot deployment on unobserved geometries. By designing the Local-TIGNN architecture to be inherently geometry-agnostic, the learning paradigm is shifted from global geometric mappings to localized interaction laws. Rather than memorizing the global mesh topology of a specific solid or fluid domain, the network encapsulates the fundamental thermodynamic exchanges between neighboring nodes. This mesh independence serves as a powerful structural inductive bias, enabling the system to instantiate stable, physically valid simulations for complex geometries never encountered during the offline training phase.

\subsection{Reasoning on Unobservable Variables}
A defining attribute of cognitive capability within this architecture is the capacity to infer physical information that remains hidden to external surface sensors. While the computer vision pipeline captures exclusively the observable boundaries—such as the free surface of a sloshing fluid or the external displacement field of a deformable solid—the Local-TIGNN leverages these sparse observations to reconstruct the complete volumetric thermodynamic state of the system. This inverse inference task enables the reconstruction of unobservable latent fields, including internal pre-stress tensors $\bs \sigma$ and localized energy density distributions $e$.

Crucially, this reconstruction process is mathematically safeguarded by the GENERIC formalism. While state-of-the-art data-driven macro-models, such as GraphCast \cite{lam2023learning}, exhibit emergent physical plausibility within their latent layers, they lack explicit structural guarantees. In contrast, the proposed architecture ensures that the inference of unobservable variables is strictly bounded by localized energy conservation laws and entropy inequality constraints. Consequently, the internal reasoning performed by the digital twin is protected against unphysical artifacts or statistical drift, guaranteeing strict compliance with the fundamental laws of thermodynamics.

\subsection{Bidirectional Flow and Decision Support}
The CDT operates within a closed-loop ecosystem characterized by a continuous, bidirectional prediction-correction cycle. In the physical-to-virtual direction, the visual perception pipeline continuously monitors the boundary configurations of the physical asset, utilizing data assimilation techniques to correct accumulated numerical integration drift and anchor the simulation to reality. Conversely, in the virtual-to-physical direction, the inferred internal fields and autoregressive rollouts are projected back onto the physical scene via Augmented Reality (AR). This bidirectional coupling transforms the CDT into an interactive decision-support tool. By rendering invisible stress fields and dissipation patterns visible to human operators, the system achieves a state of perceptual augmentation, facilitating physics-informed decision-making in real-time.

\section{Experiments and Results}
\label{sec:experiments}

To evaluate the generalization capabilities and computational efficiency of the proposed zero-shot framework, we validate the architecture across two fundamentally distinct physical domains: the continuum mechanics of a viscoelastic cantilevered beam and the free-surface dynamics of a viscous fluid. A comparative overview of the respective end-to-end digital twin pipelines is schematically illustrated in Fig.~\ref{fig:optimization_strategy}. 

While both scenarios share the same core Local-TIGNN physics engine and utilize an auxiliary initialization network ($\Psi_{\text{ini}}$) at $t=0$ to resolve latent fields, they differ significantly in their operational data-assimilation loops for $t>0$. Specifically, the solid mechanics pipeline relies on a sparse keypoint tracking mechanism driven by optical flow, whereas the fluid dynamics workflow requires continuous semantic boundary segmentation to overcome the absence of trackable material surface features.

\subsection{3D viscoelastic beam bending}

\subsubsection{Experimental Setup and Computational Foundation}

To ensure high-fidelity representation of the physical asset, we conducted a two-stage mechanical characterization. First, a preliminary estimation of the Young's modulus $E$ was obtained using Euler-Bernoulli beam theory to provide a rapid baseline. Second, to account for the observed large displacements and geometric nonlinearities, this baseline was refined using a high-fidelity Finite Element Method (FEM) model in Abaqus (Dassault Syst\`emes, 2026), employing a Kirchhoff-Saint Venant (KSV) hyperelastic formulation. It is well known that this model has significant limitations, particularly under compressive conditions, but in our experiments it has demonstrated a high degree of accuracy in predicting the observed phenomena.

\begin{figure}[htbp]
    \centering
    \includegraphics[scale=1.4]{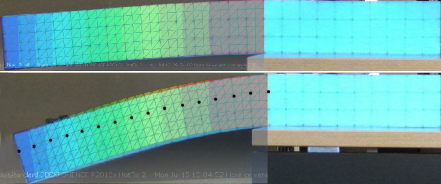} 
    \caption{\textbf{Validation of the mechanical characterization.} 
    The Finite Element Method (FEM) solution computed in Abaqus (colored mesh) is superimposed onto the experimental video frame. The spatial alignment between the numerical prediction and the physical deformation corroborates the validity of the calibrated elastic parameters and the Euler-Bernoulli beam assumption.}
    \label{fig:abaqus_match}
\end{figure}

The accuracy of this characterization was validated by superimposing the numerical simulation onto real-world video frames (Fig. \ref{fig:abaqus_match}). The spatial alignment confirms that the calibrated KSV-viscoelastic model captures the non-linear dynamics of the physical specimen, ensuring that the synthetic dataset generated for the Local-TIGNN remains grounded in the physical reality of the asset.

While large geometric transformations are handled via the KSV constitutive model,  viscoelastic stress relaxation is modeled through a single-term Prony series of the dimensionless shear relaxation modulus:
\begin{equation} \label{eq:prony}
    g_{R}(t) = 1-\bar{g}_1(1-e^{-\frac{t}{\tau_{1}}}),
\end{equation}
where $\bar{g}_1$ is the shear relaxation coefficient and $\tau_{1}$ is the relaxation time. The training scenarios were generated using $\rho = 18.44 \, \text{kg/m}^3$, $E = 4.5 \times 10^4 \, \text{Pa}$, $\nu = 0.3$, $\bar{g}_{1} = 0.5$, and $\tau_{1} = 0.05 \, \text{s}$. To induce a rich variety of bending modes, the system was subjected to distributed loads of magnitude $F \in [1, 5] \, \text{N}$, applied at varying locations and orientations normal to the surface.

\subsubsection{Data Generation and Training Protocols}

Given the hybrid nature of the Cognitive Digital Twin, two distinct datasets were curated to train the separate modules of the pipeline: the visual perception system (semantic segmentation) and the physics reasoning engine (Local-TIGNN).

\paragraph*{\textbf{Visual Perception Dataset (Real-World)}}
To train the semantic segmentation network responsible for topology initialization, a dataset of 55 real-world RGB images was collected using the laboratory camera setup. The acquisition process prioritized diversity, capturing the beam under varying lighting conditions, backgrounds, and color changes to promote robustness. Ground truth masks were generated via manual pixel-wise labeling, distinguishing between ``grid lines'' and ``background''. Given the relatively low complexity of this binary segmentation task, this compact dataset, further enriched through standard data augmentation techniques such as random rotations, brightness adjustments, and scaling, proved sufficient to ensure the network's ability to generalize to unseen experimental settings without requiring extensive data collection.

\paragraph*{\textbf{Physics Learning Dataset (Synthetic)}}
While the vision system learns from reality, the thermodynamic reasoning engine is trained on high-fidelity synthetic data to learn the underlying governing laws. Using the FEM model parameterized in the previous section (properties defined in Eq. (\ref{eq:prony})), a total of 200 simulations were executed in Abaqus.
\begin{itemize}
    \item Sampling Strategy: To cover the phase space of the beam's dynamics, random distributed loads were applied with varying magnitudes and spatial locations. Additionally, the geometric dimensions of the beam were varied, with the length ranging from 35 to 80~cm, and both width and depth spanning from 10 to 40~cm.  
    \item Data Structure: From these simulations, the nodal kinematic data (positions, velocities) and thermodynamic state variables (stress tensors, energy densities) were extracted at discrete time steps $\Delta t$. This synthetic dataset serves as the ground truth for the Local-TIGNN, enabling it to learn the causal relationships between local kinematic configurations and energy evolution without noise interference.
    \item Mesh and Spatial Discretization: A fine mesh size of 2.5~cm was utilized within the FEM solver to guarantee numerical convergence, as a coarser 5~cm mesh configuration failed to converge. Subsequently, to construct the final dataset and maintain strict spatial compatibility with the fiducial grid marked on the real physical beam, the simulated nodal data was sub-sampled by extracting state variables at 5~cm intervals.
\end{itemize}

\subsubsection{Visual Perception Performance}
\label{subsubsec:perception_performance}

The perception module provides the geometric foundation for the physics engine. To ensure that segmentation errors do not propagate as erroneous boundary conditions, the U-Net architecture is first validated on a held-out test set. The Intersection over Union (IoU) metric is employed to quantify mask overlap:
\begin{equation}
    \text{IoU} = \frac{|M_{\text{pred}} \cap M_{\text{gt}}|}{|M_{\text{pred}} \cup M_{\text{gt}}|}.
\end{equation}

The model achieved an average IoU of $0.769$. While this value indicates a moderate overlap, qualitative analysis confirms that discrepancies are primarily morphological rather than topological. The prediction error is concentrated at the boundaries of the thin grid lines. The dominance of True Positive regions confirms that the network successfully captures the structural skeleton of the specimen. Minor deviations appear as slight boundary over-segmentation, whereas False Negatives are localized and minimal. Crucially, the topology is perfectly preserved: all node intersections are detected with high spatial consistency. Since the physics engine relies on the centroids of these intersections rather than pixel-perfect widths, this IoU is sufficiently robust for downstream tracking.

\subsubsection{Physics-Informed State Initialization and Inference}
\label{subsubsec:physics_accuracy}

The physics inference framework for the structural continuum operates via a two-stage sequential pipeline: a latent state initialization stage at $t=0$ followed by the temporal rollout of the physics engine. First, because the internal pre-stress fields induced by gravity and self-weight remain completely unobservable to the monocular vision system, they must be inferred solely from the initial perceived boundary geometry $\mathcal{G}_0$. This inverse mapping is resolved by the auxiliary initialization Graph Neural Network ($\Psi_{\text{ini}}$).

Evaluated on the unseen test set, this static module successfully reconstructs the latent mechanical state, achieving a global von Mises stress root-mean-square error (RMSE) of $5.48$ MPa. Crucially, for the longitudinal normal stress component ($\sigma_{11}$)—which governs the dominant bending mechanics of the cantilever structure—the network reports an RMSE of $40.28$, translating to a low relative error of just $3.25\%$. This high fidelity in the primary stress axis ensures that $\Psi_{\text{ini}}$ accurately anchors the high-magnitude stress concentrations near the fixed boundary, eliminating numerical startup transients and initializing the downstream simulation in static equilibrium.

With the initial state successfully established, we evaluate the predictive accuracy of the autoregressive Local-TIGNN physics engine during the temporal rollout phase ($t>0$). The model was trained on 180 viscoelastic simulations and validated on 20 unseen trajectories. Table~\ref{tab:offline_metrics} reports the root-mean-square error (RMSE) and relative RRMSE for the temporal evolution of the state variables: nodal positions $\bs{q}$, velocities $\bs{v}$, and the complete stress tensor $\bs{\sigma}$.

\begin{table}[htbp]
    \centering
    \renewcommand{\arraystretch}{1.2}
    \caption{Offline accuracy metrics for the {Local-TIGNN} on the test dataset.}
    \label{tab:offline_metrics}
    \begin{tabular}{lccc}
        \toprule
        \textbf{Metric} & \textbf{Position} ($\bs{q}$) & \textbf{Velocity} ($\bs{v}$) & \textbf{Stress} ($\bs{\sigma}$) \\
        \midrule
        RMSE [SI] & $4.75 \times 10^{-4}$ & $5.26 \times 10^{-5}$ & $9.87 \times 10^{0}$ \\
        RRMSE (\%) & 0.30 & 3.80 & 11.34 \\
        \bottomrule
    \end{tabular}
\end{table}

The model exhibits high kinematic fidelity, with a sub-percent RRMSE in position ($0.30\%$) and a robust error margin in velocity ($3.80\%$). This low kinematic error demonstrates that the Local-TIGNN successfully learns a stable time-integration scheme, preventing numerical drift and preserving physical trajectory consistency between visual frames.

More importantly, the inference of the stress tensor $\bs{\sigma}$ demonstrates the model’s ability to approximate the material’s constitutive law. Although the RRMSE is higher ($11.34\%$), this is expected given the non-linear dependency of the stress field on the deformation history (defined by the Prony series). Unlike kinematic variables, which are directly constrained by the mesh geometry, the stress field is an emergent property of the internal dissipative dynamics. An $11\%$ error signifies that the network has successfully captured the viscoelastic characteristic time-scales, enabling the inference of internal stress distributions that are essential for structural health monitoring.

\subsubsection{Sim-to-Real Data Assimilation and Closed-Loop Stability}
\label{subsubsec:sim_to_real_gap}

\begin{figure*}[htbp]
    \centering
    \includegraphics[width=\textwidth]{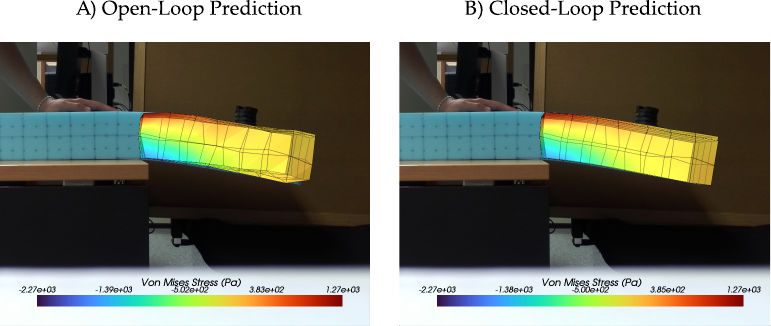} 
    \caption {Compensation of Sim-to-Real Drift. (A) Open-Loop Prediction: Despite high offline accuracy, the uncorrected simulation suffers from error accumulation due to sensor noise and material mismatch, leading to visible deformation drift.
    (B) Closed-Loop Prediction: The visual feedback loop corrects these small errors in real-time, maintaining high fidelity in both geometric alignment and the resulting stress distribution.}
    \label{fig:correction_comparison}
\end{figure*}

To validate the necessity of the visual feedback loop, we evaluate the system's resilience to the sim-to-real gap by comparing an open-loop rollout against the proposed closed-loop digital twin (Fig.~\ref{fig:correction_comparison}). Although the {Local-TIGNN} exhibits high offline accuracy on synthetic tests ($0.30\%$ positioning error), its precision degrades when deployed open-loop on the physical setup (Fig.~\ref{fig:correction_comparison}A). This divergence is a characteristic bottleneck of autoregressive architectures; unstructured real-world noise—such as camera calibration jitter and slight material parameter mismatches (e.g., stiffness and damping discrepancies)—compounds over consecutive integration steps, resulting in macroscopic geometric drift.

The closed-loop architecture successfully mitigates this error accumulation via real-time data assimilation (Fig.~\ref{fig:correction_comparison}B). By continuously injecting the tracked nodal positions into the physics engine, the feedback loop effectively resets the integration error at each time-step. This continuous geometric constraint forces the virtual mesh to remain strictly aligned with the physical specimen. Crucially, this geometric correction guarantees that the inferred latent stress fields ($\boldsymbol{\sigma}$) remain physically anchored to the actual deformations observed, preventing unphysical state estimations.

\subsubsection{Computational Efficiency and Real-Time Latency}
\label{subsec:computational_performance}
\begin{figure}
    \centering
    \includegraphics[scale=0.7]{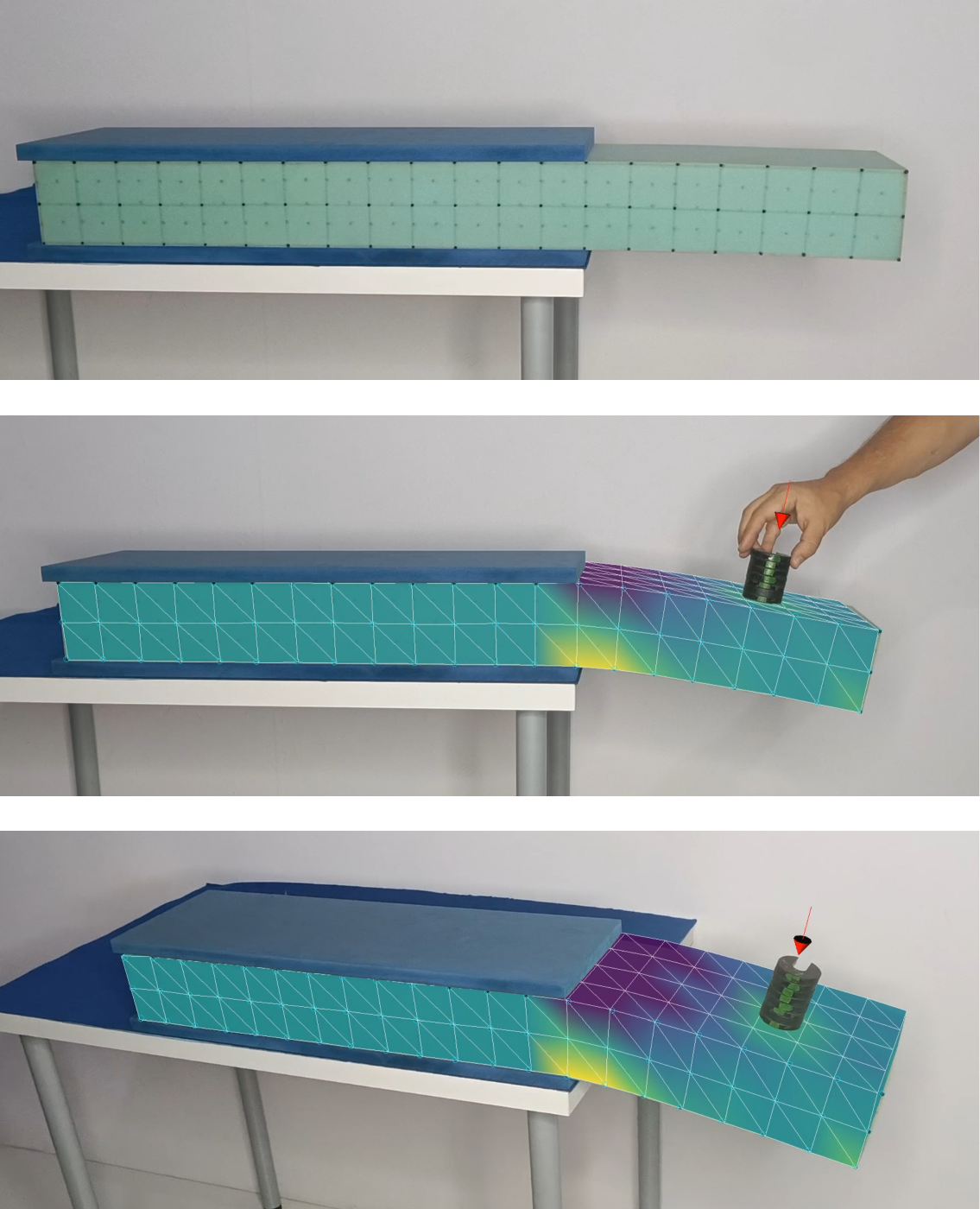}     
    \caption[Representative Operation of the Digital Twin]{
    Representative operation of the Cognitive Digital Twin on the hyperelastic beam. The top panel shows the raw physical setup, highlighting the fiducial grid used for Lagrangian tracking. The middle and bottom panels illustrate the real-time synchronization between the experimental visual data and the Local-TIGNN simulation. By projecting the inferred stress fields and mesh topology back onto the physical world, this sequence showcases the transition from raw perception to augmented intelligence, maintaining thermodynamic consistency throughout the large-deformation test.Documentary video from \protect\url{https://youtu.be/0G6n2N4OPXU?si=AZB7f884NK-_VBCi}}
    \label{fig:beam_final_representative}
\end{figure}

To confirm the operational viability of the cognitive digital twin for live industrial monitoring, we benchmark the computational latency of each pipeline component. The execution is decoupled into two phases:

\begin{itemize}
    \item \textbf{Offline Initialization:} The initial grid segmentation via the U-Net requires $0.13\,\text{s}$. Because this operation is executed only on the first frame to bootstrap the nodal graph topology, it represents a negligible one-time cost that does not constrain the dynamic loop.
    \item \textbf{Online Execution Loop:} During continuous tracking, the total per-frame latency ($\Delta t_{\text{total}} \approx 9.1\,\text{ms}$) is dominated by the sequential combination of visual tracking via sparse optical flow ($\approx 0.3\,\text{ms}$) and Local-TIGNN forward inference ($\approx 8.8\,\text{ms}$).
\end{itemize}

As demonstrated in the full synchronized sequence (Fig.~\ref{fig:beam_final_representative}), the entire perception-to-inference loop operates well within the standard real-time visualization budget of $33.3\,\text{ms}$ ($30\,\text{fps}$). With a total cycle time under $10\,\text{ms}$, the proposed framework is theoretically capable of operating at frequencies exceeding $100\,\text{Hz}$, leaving an ample safety margin for downstream decision-making, predictive control, or augmented reality rendering.

\subsection{Fluid Sloshing in Unseen Containers}
\label{subsection:fluid_sloshing}

The second case study validates the framework under free-surface fluid dynamics, specifically focusing on the sloshing behavior of a viscous fluid within containers of varying geometries. Unlike the solid mechanics scenario where individual Lagrangian nodes are tracked via a physical grid, this setup introduces highly non-linear fluid flows and moving boundaries, demanding a distinct approach to data assimilation and perception.

\subsubsection{Experimental Setup and Computational Foundation}
\label{subsubsec:fluid_dataset}

Unlike the polymer beam, which required inverse calibration to resolve material uncertainties, the fluid is well-characterized as bi-distilled glycerin at $30^\circ\text{C}$. Hydrodynamic behavior was modeled using a linear $U_s-U_p$ Hugoniot form Equation of State (EOS). The physical properties were fixed to a density of $\rho = 1261.0\,\text{kg}/\text{m}^3$, a dynamic viscosity of $\mu = 1.412\,\text{Pa}\cdot\text{s}$, and a reference sound speed of $c_0 = 13.0\,\text{m}/\text{s}$.
\subsubsection{Data Generation and Training Protocols}
\label{subsubsec:fluid_data}

Given the hybrid nature of the Cognitive Digital Twin, two distinct datasets were curated to train the separate modules of the pipeline: the visual perception system (semantic segmentation) and the physics reasoning engine (Local-TIGNN).

\paragraph*{\textbf{Visual Perception Dataset (Real-World)}}
To enable the vision pipeline to robustly isolate the fluid domain across different environments, a real-world video dataset encompassing diverse container profiles and liquid levels was curated. A total of 160 representative frames were manually annotated with pixel-level masks defining the container and fluid boundaries. To combat manual annotation sparsity and enhance the robustness of the segmentation network against environmental noise, extensive data augmentation techniques were applied, including random rotations, spatial scaling, and photometric lighting variations. This strategy allowed the semantic segmentation architecture to achieve excellent boundary generalization without requiring a prohibitively large manual labeling campaign.

\paragraph*{\textbf{Physics Learning Dataset (Synthetic)}}
The underlying fluid dynamics were learned from a synthetic corpus generated via high-fidelity Smooth Particle Hydrodynamics simulations in Abaqus. To comprehensively cover the operational space, a parametric sweep comprising 120 distinct simulations was executed by systematically varying the system across three principal axes. First, the container geometry incorporated multiple glass profiles and non-trivial cross-sections. Second, the fill ratio was modulated across varying initial fluid volumes to alter the system's natural frequencies. Lastly, the initial kinematics included distinct initial velocity and acceleration vectors applied to the vessel to trigger complex, highly non-linear sloshing modes. The resulting synthetic trajectories were split into distinct subsets to train the two core physics modules: the state-initialization network ($\Psi_{\text{ini}}$) and the Local-TIGNN engine.

\subsubsection{Visual Perception Performance}
\label{subsubsec:perception_performance_glass}

To extract the boundaries of both the fluid volume and the moving container, we employ the YOLOv11s-seg architecture, a state-of-the-art framework optimized for real-time instance segmentation. That particular model was specifically selected to balance semantic feature extraction capacity with high-frequency inference speed, utilizing a streamlined backbone enhanced with multi-scale features for precise boundary delineation. Leveraging transfer learning from a pre-trained checkpoint, the network was fine-tuned on the custom translucent fluid and glass container domain using an input resolution of $768 \times 768$ pixels for 300 epochs. The optimized parameter space, coupled with the data augmentation pipeline, enabled rapid training convergence while ensuring high generalization capabilities from a compact set of manual annotations.

Quantitative evaluation on a held-out test set composed of 15 unseen images yielded high-fidelity tracking metrics, achieving an Intersection over Union (IoU) of $0.9304$ for the fluid class and $0.9533$ for the container class. These robust metrics confirm the model's capacity to accurately isolate the dynamic, non-linear air-fluid interface, even when subjected to the severe optical refraction artifacts and variable lighting conditions inherent to transparent glassware. Crucially for the digital twin pipeline, the model demonstrated a mean inference latency of $9.8\,\text{ms}$ per frame. This sub-centisecond processing time is orders of magnitude lower than the characteristic time-scales of the fluid sloshing dynamics, ensuring that the visual perception module can operate continuously in real time without introducing synchronization bottlenecks or computational overhead to the data assimilation loop.

\subsubsection{Physics-Informed State Initialization and Inference}
\label{subsubsec:offline_physics_accuracy_glycerin}

The physics inference framework operates via a two-stage sequential pipeline: a latent state initialization phase followed by an autoregressive temporal rollout of the fluid continuum. First, because the unobserved initial velocity and internal energy fields remain latent to the vision system, they must be reconstructed from the perceived initial boundary geometry and the container's kinematic excitation vector. This task is executed by a lightweight auxiliary Graph Neural Network (GNN) acting as a static encoder-decoder without temporal recurrence. Comprising two hidden layers of 80 units (172k trainable parameters), this initialization module successfully ``cold-starts'' the simulation, achieving velocity root-mean-square errors (RMSE) on the test set of $0.027\,\text{m/s}$ ($v_x$), $0.083\,\text{m/s}$ ($v_y$), and $0.034\,\text{m/s}$ ($v_z$). The marginally elevated discrepancy in the vertical component ($v_y$) is physically consistent with the primary axis of inertial sloshing excitation during the container's sudden deceleration. Crucially, the internal energy density is reconstructed with an RMSE of $4.53 \times 10^{-8}$, anchoring the simulation within a thermodynamically consistent state prior to dynamic rollout.

Once initialized, the Local-TIGNN architecture governs the forward temporal prediction of the fluid. To preserve inductive structural biases while retaining high learning capacity, the network features two hidden layers of 100 units and executes seven message-passing steps per time-step (299k trainable parameters). To enforce strict adherence to thermodynamic conservation laws, the Lagrangian constraint weight was regularized with a penalty of $\lambda_d = 50$, and an inductive noise variance of $8 \times 10^{-3}$ was injected during training to enhance robustness against autoregressive drift. Convergence was accelerated via a warm-start transfer learning strategy, initializing the network with weights optimized on a foundational water sloshing baseline and subsequently fine-tuning on the specific bi-distilled glycerin dataset. The choice of glycerin isolates a highly viscous regime where the dampening of high-frequency spatial turbulence significantly stabilizes the downstream vision-tracking and data-assimilation tasks. 

\begin{table}[htbp]
    \centering
    \renewcommand{\arraystretch}{1.2}
    \caption{Quantitative performance metrics for the Local-TIGNN on the unseen glycerin test dataset.}
    \label{tab:glycerin_metrics}
    \begin{tabular}{lccc}
        \toprule
        \textbf{Metric} & \textbf{Position} ($\bs{q}$) & \textbf{Velocity} ($\bs{v}$) & \textbf{Energy} ($\bs{e}$) \\
        \midrule
        RMSE [SI] & $1.50 \times 10^{-3}$ & $8.70 \times 10^{-3}$ & $4.53 \times 10^{-8}$ \\
        RRMSE (\%) & 2.49 & 26.34 & 0.005 \\
        \bottomrule
    \end{tabular}
\end{table}

Table~\ref{tab:glycerin_metrics} outlines the quantitative accuracy of the Local-TIGNN across unseen test trajectories. The model exhibits high geometric tracking performance, yielding a positional RMSE of $1.50 \times 10^{-3}\,\text{m}$ (a relative error of $2.49\%$), significantly outperforming prior ungrounded water baselines. While the absolute velocity error remains tightly bounded at $8.70 \times 10^{-3}\,\text{m/s}$, its corresponding relative metric (RRMSE) shows a nominal inflation to $26.34\%$. This phenomenon is purely a numerical artifact of the relative calculation rather than an indicator of physical degradation; owing to glycerin's high viscosity, the fluid spends extensive intervals in near-equilibrium or fully dissipated states where the ground-truth velocity approaches zero, thus inflating the percentage error due to a vanishingly small denominator. Furthermore, the internal energy density registers a negligible relative error of $0.005\%$, corroborating the thermodynamic rigor of the network.

From a computational standpoint, the entire physics rollout pipeline demonstrates a mean inference latency of $13.6\,\text{ms}$ per frame. This execution budget is split between the neural network forward pass ($10.23\,\text{ms}$) and the dynamic re-computation of the mesh graph topology ($3.4\,\text{ms}$). This sub-frame latency comfortably satisfies real-time execution constraints, preventing phase lags during the closed-loop data assimilation loop.

\subsubsection{Sim-to-Real Data Assimilation and Closed-Loop Stability}
\label{subsubsec:sloshing_qualitative}

\begin{figure*}[htbp]
    \centering
    \includegraphics[width=\textwidth]{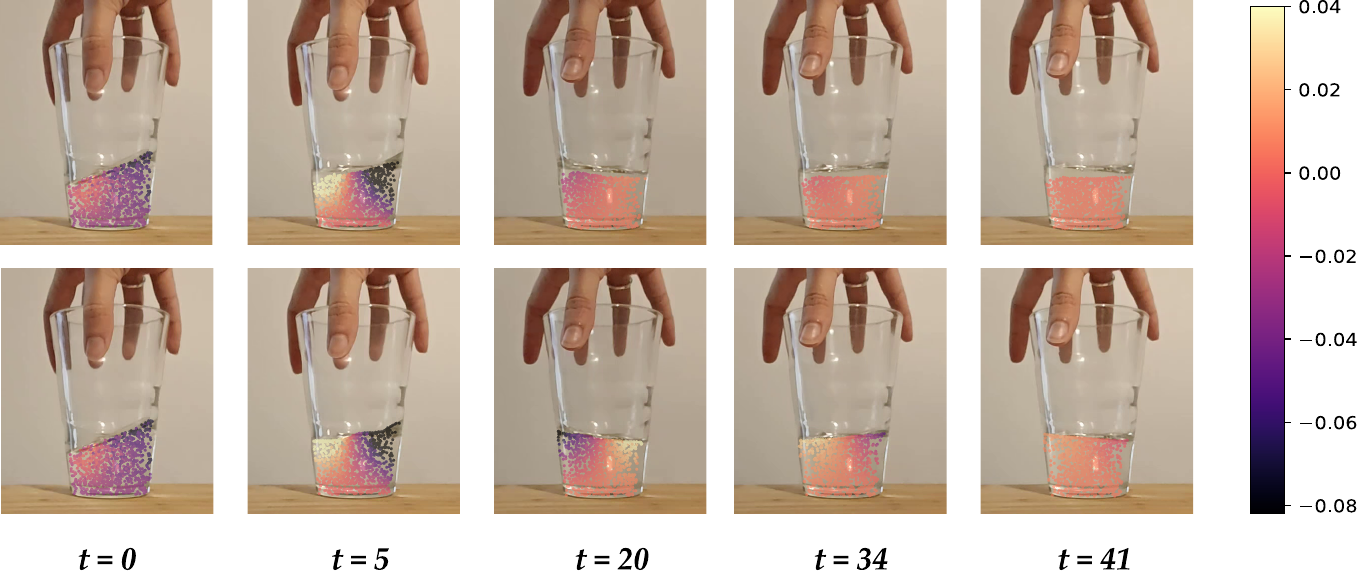} 
    \caption{Dynamic synchronization and latent vertical velocity field ($v_y$) inference. Temporal evolution of the sloshing prediction from $t=0$ to $t=41$, where particles are colored by their vertical velocity component. The open-loop rollout (top) suffers from progressive phase drift and premature numerical damping, incorrectly predicting hydrostatic rest by $t=34$ while the physical fluid is still in motion. The closed-loop architecture (bottom) leverages visual feedback to maintain strict dynamic consistency, preserving active velocity gradients at $t=41$ that correspond to the actual physical oscillations.Documentary video from \protect\url{https://youtu.be/-r5nOTeaUvQ?si=MO_SmqJS4wtdy87F}.}
    \label{fig:sloshing_velocity_comparison}
\end{figure*}

To validate the real-world deployment of the fluid digital twin and its capability to maintain long-term dynamic synchronization, we evaluate the inferred latent fields against the observed macroscopic fluid motion. Free-surface sloshing exhibits severe sensitivity to initial conditions and non-linear boundary damping, making dense experimental tracking of velocity or internal energy vectors unfeasible in a live setting. As illustrated in the top rollout of Fig.~\ref{fig:sloshing_velocity_comparison}, the open-loop prediction rapidly succumbs to accumulative errors. Beyond significant phase drift, the uncorrected simulation suffers from premature kinetic damping; by time-step $t=34$, the open-loop rollout settles into an unphysical hydrostatic equilibrium with vanishingly small velocities, whereas the physical fluid remains actively oscillating. This divergence indicates that the sim-to-real gap is driven not only by parametric mismatches (e.g., idealized viscosity or wall-friction formulations), but by the cascading propagation of minor errors across the segmentation and initialization stages during autoregressive integration.

In contrast, the closed-loop architecture successfully bridges this gap by assimilating the visual segmentation masks as continuous boundary constraints (Fig.~\ref{fig:sloshing_velocity_comparison}, bottom row). This real-time feedback loop goes beyond surface matching; it implicitly adjusts the underlying kinetic energy state. By constraining the fluid domain to conform to the observed free-surface profile, the network is compelled to induce non-zero vertical velocity ($v_y$) gradients that mirror the residual physical oscillations. Consequently, the digital twin yields a physically anchored visualization of the latent flow fields, capturing wave dynamics that vanish entirely in ungrounded simulations.

\subsubsection{Computational Efficiency and Real-Time Latency}
\label{subsubsec:computational_time}

To guarantee the operational feasibility of the cognitive fluid digital twin for interactive applications, the end-to-end latency of the closed-loop pipeline is benchmarked. The total execution time per frame ($\Delta t_{\text{total}}$) is modeled as the sequential sum of the semantic segmentation inference ($\Delta t_{\text{vis}}$), the forward physics rollout including dynamic graph re-meshing ($\Delta t_{\text{phys}}$), and the geometric volume-assimilation step ($\Delta t_{\text{sync}}$). 

The visual perception module processing the YOLOv11s-seg network requires $\Delta t_{\text{vis}} = 9.80\,\text{ms}$, while the Local-TIGNN core demands $\Delta t_{\text{phys}} = 13.60\,\text{ms}$ to execute the forward neural pass and re-compute the mesh graph topology. Incorporating the column-wise geometric correction step ($\Delta t_{\text{sync}} = 1.83\,\text{ms}$), the framework achieves a total per-frame processing latency of approximately $25.23\,\text{ms}$, translating to an operational throughput of $\approx 39.6\,\text{Hz}$. 

In the context of human-computer interaction and online monitoring, the standard real-time visualization threshold is established at $30\,\text{Hz}$ ($\approx 33.3\,\text{ms}$). Operating at $25.23\,\text{ms}$, the proposed pipeline runs comfortably below this computational budget. This unallocated overhead ensures that the architecture can support downstream real-time rendering pipelines, edge-computing communication protocols, or predictive control algorithms without introducing synchronization bottlenecks.

\section{Conclusion and Future Work}
\label{sec:conclusion}

This work introduces a unified framework that bridges computer vision and continuum mechanics through a real-time, closed-loop data assimilation pipeline. By mapping raw visual tracking data into dynamic graph representations processed by a Local-TIGNN engine, the proposed architecture moves beyond mere geometric localization to infer latent, unobservable thermodynamic state variables, such as internal pre-stress tensors, energy density distributions, and full-field velocity vectors. Enforcing the GENERIC formalism directly within the neural architecture guarantees that these latent estimations remain strictly consistent with thermodynamic invariants. This constraints-by-design approach provides a physically grounded representation of internal dynamics that are otherwise inaccessible to standard optical sensors alone.

A key attribute of the developed methodology is its capacity for zero-shot deployment on novel geometries and boundary conditions without requiring case-specific retraining. Because the underlying physical principles are encapsulated as localized interaction laws rather than global spatial mappings, the geometry-agnostic framework seamlessly adapts to distinct physical regimes. This multi-domain flexibility has been validated across both the structural mechanics of a viscoelastic cantilever beam and the non-linear hydrodynamics of fluid sloshing using a unified graph-based solver. The experimental results demonstrate that the core physics engine generalizes effectively across disparate setups, provided the upstream perception module delivers appropriate geometric boundaries.

The operational viability of the framework for online monitoring and interactive applications is corroborated by its high computational efficiency. Achieving end-to-end processing latencies well below the $33.3\,\text{ms}$ real-time threshold—ranging from $9.1\,\text{ms}$ ($\approx 110\,\text{Hz}$) for structural systems to $25.23\,\text{ms}$ ($\approx 39.6\,\text{Hz}$) for complex fluid domains—the pipeline runs comfortably within standard visualization budgets. Furthermore, embedding thermodynamic inductive biases serves as a robust physical regularizer, stabilizing the autoregressive rollouts against the unstructured noise and pixel jitter inherent in real-world sensory inputs, thereby ensuring long-term tracking stability.

Despite its robust performance, the current implementation is bounded by a planar motion assumption, making the perception module sensitive to significant out-of-plane displacements along the depth axis, which can introduce projection errors during graph initialization. However, due to the modular design of the architecture, this limitation offers a clear path for subsequent developments. Future research will focus on integrating depth-sensing modalities, such as RGB-D cameras or stereoscopic vision systems. Incorporating volumetric perception will enhance the framework's geometric robustness, extending the capabilities of the digital twin to fully unconstrained three-dimensional dynamics while preserving the underlying physics-informed reasoning engine. Ultimately, this paradigm opens new avenues for cognitive digital twins, paving the way toward advanced perceptual augmentation and interactive decision-support systems in robotic and industrial environments.

\section*{Acknowledgments}

The authors acknowledge the support of the Spanish Ministry of Science and Innovation, AEI/10.13039/501100011033 (grant no. PID2023-147373OB-I00). The authors also acknowledge the support of the Ministry for Digital Transformation and the Civil Service, through the ENIA 2022 Chairs for the creation of the university-industry chairs in Artificial Intelligence, through grant TSI-100930-2023-1.

Funded by the European Union. Views and opinions expressed are however those of the author(s) only and do not necessarily reflect those of the European Union or the European Research Council Executive Agency. Neither the European Union nor the granting authority can be held responsible for them.

This work is supported by ERC grant PHYSIA 101264273.

\bibliography{references} 
\bibliographystyle{unsrt}

\end{document}